\documentclass{article}
\usepackage{ismir,amsmath,cite,url}
\usepackage{graphicx}
\usepackage{color}

\usepackage{subcaption}
\usepackage[export]{adjustbox}

\title{Transfer learning for\\music classification and regression tasks}

\twoauthors
  {Keunwoo Choi, Gy\"orgy Fazekas, Mark Sandler} {Centre for Digital Music\\Queen Mary University of London, London, UK\\{\tt keunwoo.choi@qmul.ac.uk}}
  {Kyunghyun Cho} {Center for Data Science\\New York University, New York, NY, USA\\{\tt kyunghyun.cho@nyu.edu}}

\usepackage{enumitem} 

\begin{document}

\maketitle
\begin{abstract}
In this paper, we present a transfer learning approach for music classification and regression tasks. We propose to use a \textit{pre-trained convnet feature}, a concatenated feature vector using the activations of feature maps of multiple layers in a trained convolutional network. We show how this convnet feature can serve as general-purpose music representation. In the experiments, a convnet is trained for music tagging and then transferred to other music-related classification and regression tasks. The convnet feature outperforms the baseline MFCC feature in all the considered tasks and several previous approaches that are aggregating MFCCs as well as low- and high-level music features.

\end{abstract}

\vspace{-5pt}
\section{Introduction}\label{sec:introduction}
In the field of machine learning, transfer learning is often defined as \textit{re-using parameters} that are trained on a \textit{source} task for a \textit{target} task, aiming to transfer knowledge between the domains. A common motivation for transfer learning is the lack of sufficient training data in the target task. When using a neural network, by transferring pre-trained weights, the number of trainable parameters in the target-task model can be significantly reduced, enabling effective learning with a smaller dataset.

A popular example of transfer learning is semantic image segmentation in computer vision, where the network utilises rich information, such as basic shapes or prototypical templates of objects, that were captured when trained for image classification \cite{oquab2014learning}. Another example is pre-trained word embeddings in natural language processing. Word embedding, a vector representation of a word, can be trained on large datasets such as Wikipedia \cite{mikolov2013efficient} and adopted to other tasks such as sentiment analysis \cite{le2014distributed}.

There have been several works on transfer learning in Music Information Retrieval (MIR). Hamel~et~al. proposed to directly learn music features using linear embedding \cite{weston2011wsabie} of mel-spectrogram representations and genre/similarity/tag labels \cite{hamel2013transfer}. Oord~et~al. outlines a large-scale transfer learning approach, where a multi-layer perceptron is combined with the spherical K-means algorithm \cite{coates2012learning} trained on tags and play-count data \cite{van2014transfer}. After training, the weights are transferred to perform genre classification and auto-tagging with smaller datasets. In music recommendation, Choi~et~al. used the weights of a convolutional neural network for feature extraction in playlist generation \cite{choi2016towards}, while Liang~et~al. used a multi-layer perceptron for feature extraction of content-aware collaborative filtering ~\cite{liang2015content}.

\vspace{-2pt}
\section{Transfer Learning For Music}
\vspace{-2pt}
In this section, our proposed transfer learning approach is described. A convolutional neural network (convnet) is designed and trained for a source task, and then, the network with trained weights is used as a feature extractor for target tasks. The schematic of the proposed approach is illustrated in Figure \ref{figure:convnet}. 

\vspace{-4pt}
\subsection{Convolutional Neural Networks for Music Tagging}
\vspace{-2pt}

\begin{figure}[t!]
	\begin{center}
		\centerline{\includegraphics[width=0.93\columnwidth,  trim={0 0 0 0.5cm}]{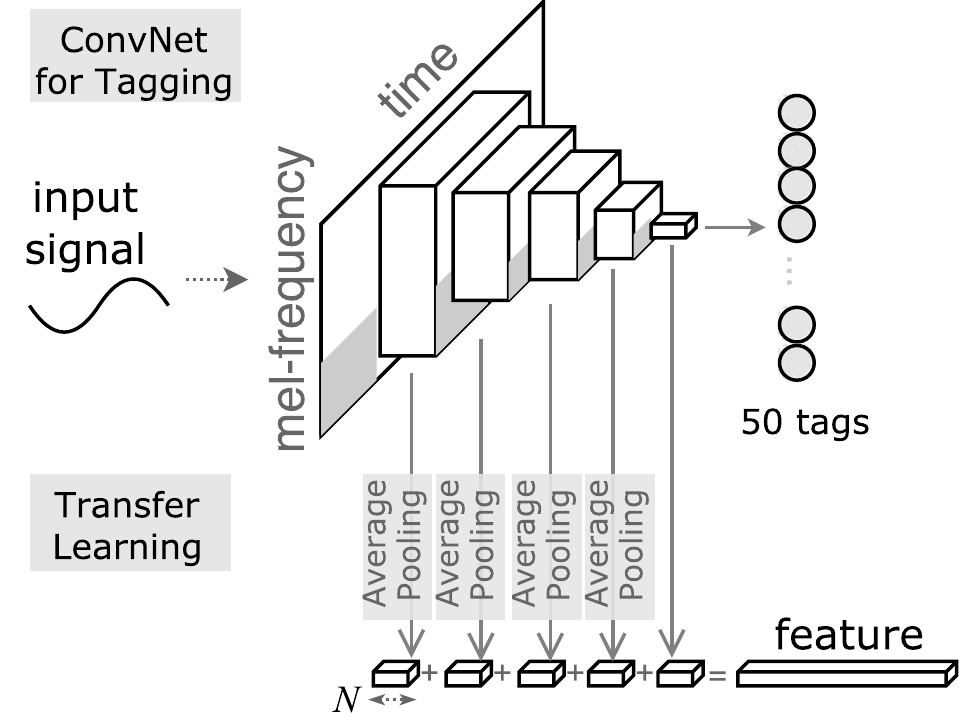}}
		\vspace{-0.15cm}
		\caption{\small A block diagram of the training and feature extraction procedures. Exponential linear unit (ELU) is used as an activation function in all convolutional layers \cite{clevert2015fast}. Max-pooling of (2, 4), (4, 4), (4, 5), (2, 4), (4, 4) is applied after every convolutional layer respectively. In all the convolutional layers, the kernel sizes are (3, 3), numbers of channels $N$ is 32, and Batch normalisation is used \cite{ioffe2015batch}. The input has a single channel, 96-mel bins, and 1360 temporal frames.
			After training, the feature maps from $1$st--$4$th layers are subsampled using average pooling while the feature map of $5$th layer is used as it is, since it is already scalar (size $1 \times 1$). Those $32$-dimensional features are concatenated to form a \textit{convnet feature}. 
		} \vspace{-0.8cm}

		\label{figure:convnet}
	\end{center}
\end{figure} 

We choose music tagging as a source task because \textit{i)} large training data is available and \textit{ii)} its rich label set covers various aspects of music, e.g., \textit{genre}, \textit{mood}, \textit{era}, and \textit{instrumentations}. In the source task, a mel-spectrogram ($\textbf{X}$), a two-dimensional representation of music signal, is used as the input to the convnet. The mel-spectrogram is selected since it is psychologically relevant and computationally efficient. It provides a mel-scaled frequency representation which is an effective approximation of human auditory perception \cite{moore2012introduction} and typically involves compressing the frequency axis of short-time Fourier transform representation (e.g., 257/513/1025 frequency bins to 64/96/128 Mel-frequency bins). In our study, the number of mel-bins is set to 96 and the magnitude of mel-spectrogram is mapped to decibel scale ($\log_{10} \textbf{X}$), following \cite{choi2016automatic} since it is also shown to be crucial in \cite{choi2017effects}.

In the proposed system, there are five layers of convolutional and sub-sampling in the convnet as shown in Figure \ref{figure:convnet}.
This convnet structure with 2-dimensional $3$$\times$$3$ kernels and 2-dimensional convolution, which is often called \textit{Vggnet} \cite{simonyan2014very}, is expected to learn hierarchical time-frequency patterns. This structure was originally proposed for visual image classification and has been found to be effective and efficient in music classification\footnote{For more recent information on kernel shapes for music classification, please see \cite{pons2017designing}.} \cite{choi2017convolutional}.

\begin{table*}[h]
	\centering
	\small
	\label{table:details_dbs}
	\begin{tabular}{l l r l l r}
		{\bf Task}                                       & {\bf Dataset name}               & \begin{tabular}[c]{@{}l@{}}\textbf{$\#$clips}\end{tabular} & \begin{tabular}[c]{@{}l@{}} \textbf{Metric}\end{tabular} & \begin{tabular}[c]{@{}l@{}}\textbf{$\#$classes} \end{tabular} \\ \hline \hline
		
		{\bf T1.} Ballroom dance genre classification        & \begin{tabular}[c]{@{}l@{}}Extended ballroom  \cite{marchand2016extended}\end{tabular}   & 4,180  & Accuracy & 13\\ \hline
		{\bf T2.} Genre classification        & \begin{tabular}[c]{@{}l@{}}Gtzan genre  \cite{tzanetakis2002musical}\end{tabular}       & 1,000    &  Accuracy & 10\\ \hline
		{\bf T3.} Speech/music classification & \begin{tabular}[c]{@{}l@{}} Gtzan speech/music  \cite{tzanetakis1999gtzan}\end{tabular} & 128       & Accuracy & 2\\ \hline
		{\bf T4.} Emotion prediction          & \begin{tabular}[c]{@{}l@{}} EmoMusic (45-second)  \cite{soleymani20131000}\end{tabular}   & 744  &   \begin{tabular}[c]{@{}l@{}}Coefficient of \\ determination ($r^2$)\end{tabular}   & \begin{tabular}[c]{@{}l@{}}N/A\\ (2-dimensional)\end{tabular} \\ \hline
		{\bf T5.} Vocal/non-vocal classification             & \begin{tabular}[c]{@{}l@{}}Jamendo  \cite{bibA03} \end{tabular}   & 4,086       & Accuracy & 2         \\ \hline
		{\bf T6.} Audio event classification    & \begin{tabular}[c]{@{}l@{}}Urbansound8K  \cite{Salamon:UrbanSound:ACMMM:14}\end{tabular}       & 8,732  & Accuracy  & 10
	\end{tabular}
	\vspace{-0.25cm}
	\caption{The details of the six tasks and datasets used in our transfer learning evaluation.}
	\label{table:details}
\end{table*}

\vspace{-6pt}
\subsection{Representation Transfer}
\vspace{-2pt}
In this section, we explain how features are extracted from a pre-trained convolutional network. In the remainder of the paper, this feature is referred to as \textit{pre-trained convnet feature}, or simply \textit{convnet feature}. 

It is already well understood how deep convnets learn \textit{hierarchical features} in visual image classification \cite{zeiler2014visualizing}. By convolution operations in the forward path, lower-level features are used to construct higher-level features. Subsampling layers reduce the size of the feature maps while adding local invariance. In a deeper layer, as a result, the features become more invariant to (scaling/location) distortions and more relevant to the target task. 

This type of hierarchy also exists when a convnet is trained for a music-related task. Visualisation and sonification of convnet features for music genre classification has shown the different levels of hierarchy in convolutional layers \cite{choi2015auralisation}, \cite{choi2016explaining}.

Such a hierarchy serves as a motivation for the proposed transfer learning. Relying solely on the last hidden layer may not maximally extract the knowledge from a pre-trained network. For example, low-level information such as tempo, pitch, (local) harmony or envelop can be captured in early layers, but may not be preserved in deeper layers due to the constraints that are introduced by the network structure: aggregating local information by discarding less-relevant information in subsampling. For the same reason, deep scattering networks \cite{bruna2013invariant} and a convnet for music tagging introduced in \cite{lee2017multi} use multi-layer representations.

Based on this insight, we propose to use not only the activations of the final hidden layer but also the activations of (up to) \textit{all} intermediate layers to find the most effective representation for each task. The final feature is generated by concatenating these features as demonstrated in Figure \ref{figure:convnet}, where all the five layers are concatenated to serve as an example. 

Given five layers, there are $\sum_{n=1}^{5} {}_5 C_n=31$ strategies of layer-wise combination. In our experiment, we perform a nearly exhaustive search and report all results. We designate each strategy by the indices of layers employed. For example, a strategy named `\texttt{135}' refers to using a $32$~$\times$~$3$~$=$~$96$-dimensional feature vector that concatenates the first, third, and fifth layer convnet features.

During the transfer, average-pooling is used for the 1st--4th layers to reduce the size of feature maps to $1$$\times$$1$ as illustrated in Figure \ref{figure:convnet}. Averaging is chosen instead of \texttt{max} pooling because it is more suitable for summarising the global statistics of large regions, as done in the last layer in \cite{lin2013network}. Max-pooling is often more suitable for capturing the existence of certain patterns, usually in small and local regions\footnote{Since the average is affected by zero-padding which is applied to signals that are shorter than 29 seconds, those signals are repeated to create 29-second signals. This only happens in Task 5 and 6 in the experiment.}.

Lastly, there have been works suggesting random-weights (deep) neural networks including deep convnet can work well as a feature extractor \cite{huang2004extreme} \cite{zeng2015traffic} (Not identical, but a similar approach is transferring knowledge from an irrelevant domain, e.g., visual image recognition, to music task \cite{gwardys2014deep}.) We report these results from random convnet features and denote it as \textit{random convnet feature}. Assessing performances of random convnet feature will help to clarify the contributions of the pre-trained knowledge transfer versus the contributions of the convnet structure and nonlinear high-dimensional transformation.

\vspace{-6pt}
\subsection{Classifiers and Regressors of Target Tasks}
\vspace{-4pt}

Variants of support vector machines (SVMs) \cite{suykens1999least, smola2004tutorial} are used as a classifier and regressor. SVMs work efficiently in target tasks with small training sets, and outperformed K-nearest neighbours in our work for all the tasks in a preliminary experiment. Since there are many works that use hand-written features and SVMs, using SVMs enables us to focus on comparing the performances of features.

\vspace{-5pt}
\section{Preparation}

\subsection{Source Task: Music Tagging} \label{subsec:original_task}
In the source task, 244,224 preview clips of the Million Song Dataset \cite{bertin2011million} are used (201,680/12,605/25,940 for training/validation/test sets respectively) with top-50 \textit{last.fm} tags including genres, eras, instrumentations, and moods. Mel-spectrograms are extracted from music signals in real-time on the GPU using \textit{Kapre} \cite{choi2017kapre}. Binary cross-entropy is used as the loss function during training. 
The ADAM optimisation algorithm \cite{kingma2014adam} is used for accelerating stochastic gradient descent. The convnet achieves 0.849 AUC-ROC score (Area Under Curve - Receiver Operating Characteristic) on the test set. We use the \textit{Keras} \cite{chollet2015} and \textit{Theano} \cite{team2016theano} frameworks in our implementation. 

\subsection{Target Tasks}
Six datasets are selected to be used in six target tasks. They are summarised in Table \ref{table:details}.
\vspace{-0.2cm}
\begin{itemize}[leftmargin=*]
	\item Task 1: The Extended ballroom dataset consists of specific Ballroom dance sub-genres.
	\vspace{-0.2cm}
	\item Task 2: The Gtzan genre dataset has been extremely popular, although some flaws have been found \cite{sturm2013gtzan}. 
	\vspace{-0.2cm}
	\item Task 3: The dataset size is smaller than the others by an order of magnitude.
	\vspace{-0.2cm}
	\item Task 4: Emotion predition on the arousal-valence plane. We evaluate arousal and valence separately. We trim and use the first 29-second from the 45-second signals.
	\vspace{-0.2cm}
	\item Task 5. Excerpts are subsegments from tracks with binary labels (\textit{`vocal'} and \textit{`non-vocal'}). Many of them are shorter than 29s. This dataset is provided for benchmarking frame-based vocal detection while we use it as a pre-segmented classification task, which may be easier than the original task.
	\vspace{-0.2cm}
	\item Task 6: This is a non-musical task. For example, the classes include \textit{air conditioner}, \textit{car horn}, and \textit{dog bark}. All excerpts are shorter than 4 seconds. 
\end{itemize}

\vspace{-5pt}
\vspace{-5pt}
\subsection{Baseline Feature and Random Convnet Feature}
\vspace{-4pt}
As a baseline feature, the means and standard deviations of 20 Mel-Frequency Cepstral Coefficients (MFCCs), and their first and second-order derivatives are used. In this paper, this baseline feature is called \textit{MFCCs} or \textit{MFCC vectors}. MFCC is chosen since it has been adopted in many music information retrieval tasks and is known to provide a robust representation.
\textit{Librosa} \cite{brian_mcfee_2015_32193} is used for MFCC extraction and audio processing.

The random convnet feature is extracted using the identical convnet structure of the source task and after random weights initialisation with a normal distribution \cite{he2015delving} but without a training.
\vspace{-5pt}
\section{Experiments}

\subsection{Configurations} \label{subsec:config_exp}
For Tasks 1-4, the experiments are done with 10-fold cross-validation using stratified splits. For Task 5, pre-defined training/validation/test sets are used. The experiment on Task 6 is done with 10-fold cross-validation without replacement to prevent using the sub-segments from the same recordings in training and validation.
The SVM parameters are optimised using grid-search based on the validation results. Kernel type/bandwidth of radial basis function and the penalty parameter are selected from the ranges below:
\vspace{-3pt}
\begin{itemize}[leftmargin=*]
	\item Kernel type: $[$\textit{linear}, \textit{radial}$]$ 
	\vspace{-0.2cm}
	\begin{itemize}
		\item Bandwidth $\gamma$ in radial basis function : \\$[1/2^3, 1/2^5, 1/2^7, 1/2^9, 1/2^{11}, 1/2^{13}, 1/N_f]$ 
	\end{itemize}
	\vspace{-0.3cm}
	\item Penalty parameter $C$ : $[0.1, 2.0, 8.0, 32.0]$ 
\end{itemize}
\vspace{-2pt}
A radial basis function is $\exp(- \gamma | \textit{x} - \textit{x} ^\prime| ^2)$, and $\gamma$ and $N_f$ refer to the radial kernel bandwidth and the dimensionality of feature vector respectively. With larger $C$, the penalty parameter or regularisation parameter, the loss function gives more penalty to misclassified items and vice versa. 
We use \textit{Scikit-learn} \cite{scikit-learn} for these target tasks.
The code for the data preparation, experiment, and visualisation are available on GitHub\footnote{\url{https://github.com/keunwoochoi/transfer_learning_music}}.
\vspace{-4pt}
\subsection{Results and Discussion} \label{subsec:discuss}
\vspace{-4pt}
\begin{figure}[t]
	\centering
	\includegraphics[width=1.0\columnwidth]{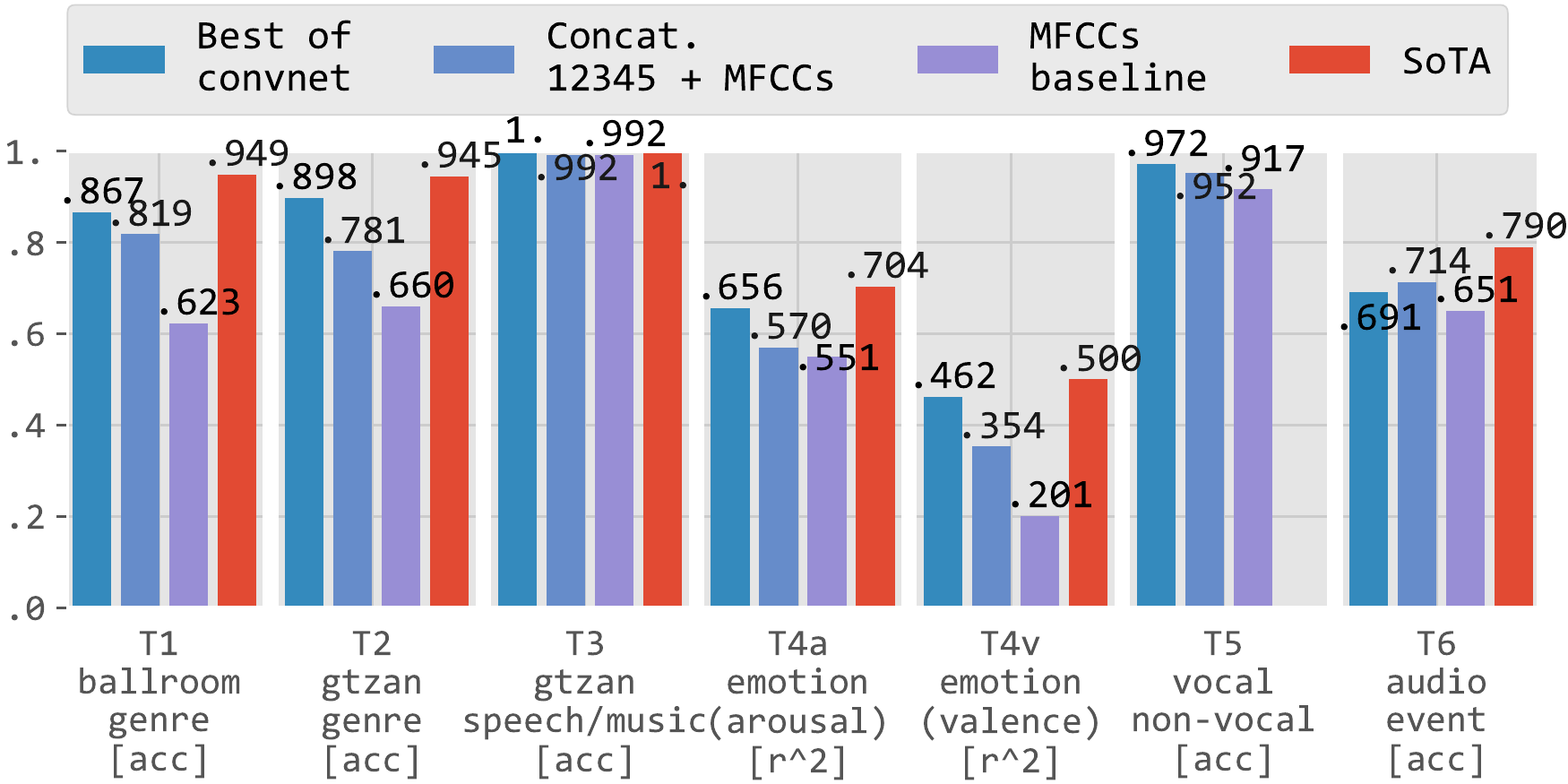}
	\vspace{-0.6cm}
	\caption{\small Summary of performances of the convnet feature (blue), MFCCs (purple), and state-of-the-art (red) for Task 1-6 (State-of-the-art of Task 5 does not exist).} 
	
	\label{fig:transferred_result}
\end{figure}

Figure \ref{fig:transferred_result} shows a summary of the results. The scores of the \textit{i)} best performing convnet feature, \textit{ii)} concatenating `\texttt{12345}'\footnote{Again, `\texttt{12345}' refers to the convnet feature that is concatenated from 1st--5th layers. For another example, `\texttt{135}' means concatenating the features from first, third, and fifth layers.} convnet feature and MFCCs, \textit{iii)} MFCC feature, and \textit{iv)} state-of-the-art algorithms for all the tasks.

In all the six tasks, the majority of convnet features outperforms the baseline feature. Concatenating MFCCs with `\texttt{12345}' convnet feature usually does not show improvement over a pure convnet feature except in Task 6, audio event classification. Although the reported state-of-the art is typically better, almost all methods rely on musical knowledge and hand-crafted features, yet our features perform competitively. An in-depth look at each task is therefore useful to provide insight.

In the following subsections, the details of each task are discussed with more results presented from (almost) exhaustive combinations of convnet features as well as random convnet features at all layers. For example, in Figure~\ref{fig:result_all_t1}, the scores of 28 different convnet feature combinations are shown with blue bars. The narrow, grey bars next to the blue bars indicate the scores of random convnet features. The other three bars on the right represent the scores of the concatenation of `\texttt{12345}' + MFCC feature, MFCC feature, and the reported state-of-the-art methods respectively. The rankings within the convnet feature combinations are also shown \textit{in} the bars where top-7 and lower-7 are highlighted.

We only briefly discuss the results of random convnet features here. The best performing random convnet features do not outperform the best-performing convnet features in any task. In most of the combinations, convnet features outperformed the corresponding random convnet features, although there are few exceptions. However, random convnet features also achieved comparable or even better scores than MFCCs, indicating \textit{i)} a significant part of the strength of convnet features comes from the network structure itself, and \textit{ii)} random convnet features can be useful especially if there is not a suitable source task.
\vspace{-2pt}
\subsubsection{Task 1. Ballroom Genre Classification}
\vspace{-2pt}
Figure \ref{fig:result_all_t1} shows the performances of different features for Ballroom dance classification. The highest score is achieved using the convnet feature `\texttt{123}' with 86.7\% of accuracy. The convnet feature shows good performances, even outperforming some previous works that explicitly use rhythmic features. 

The result clearly shows that low-level features are crucial in this task. All of the top-7 strategies of convnet feature include the \textit{second} layer, and $6/7$ of them include the \textit{first} layer. On the other hand, the lower-7 are [`\texttt{5}', `\texttt{4}', `\texttt{3}', `\texttt{45}', `\texttt{35}', `\texttt{2}', `\texttt{25}'], none of which includes the first layer. Even `\texttt{1}' achieves a reasonable performance (73.8\%).

The importance of low-level features is also supported by known properties of this task. The ballroom genre labels are closely related to rhythmic patterns and tempo \cite{marchand2016extended} \cite{sturm2016revisiting}. However, there is no label directly related to tempo in the source task. Moreover, deep layers in the proposed structure are conjectured to be mostly invariant to tempo. As a result, high-level features from the fourth and fifth layers poorly contribute to the task relative to those from the first, second, and third layers.

The state-of-the-art algorithm which is also the only algorithm that used the same dataset due to its recent release uses \textit{2D scale transform}, an alternative representation of music signals for rhythm-related tasks \cite{marchand2016scale}, and reports 94.9\% of weighted average recall.
For additional comparisons, there are several works that use the Ballroom dataset \cite{gouyon2004evaluating}. This has 8 classes and it is smaller in size than the Extended Ballroom dataset (13 classes). Laykartsis~and~Lerch \cite{lykartsis2015beat} combines beat histogram and timbre features to achieve 76.7\%. Periodicity analysis with SVM classifier in Gkiokas~et~al.~\cite{gkiokas2016towards} respectively shows 88.9\%/85.6 - 90.7\%, before and after feature selection.

\subsubsection{Task 2. Gtzan Music Genre Classification}
\begin{figure}[t]
	\centering
	\includegraphics[width=1.\columnwidth]{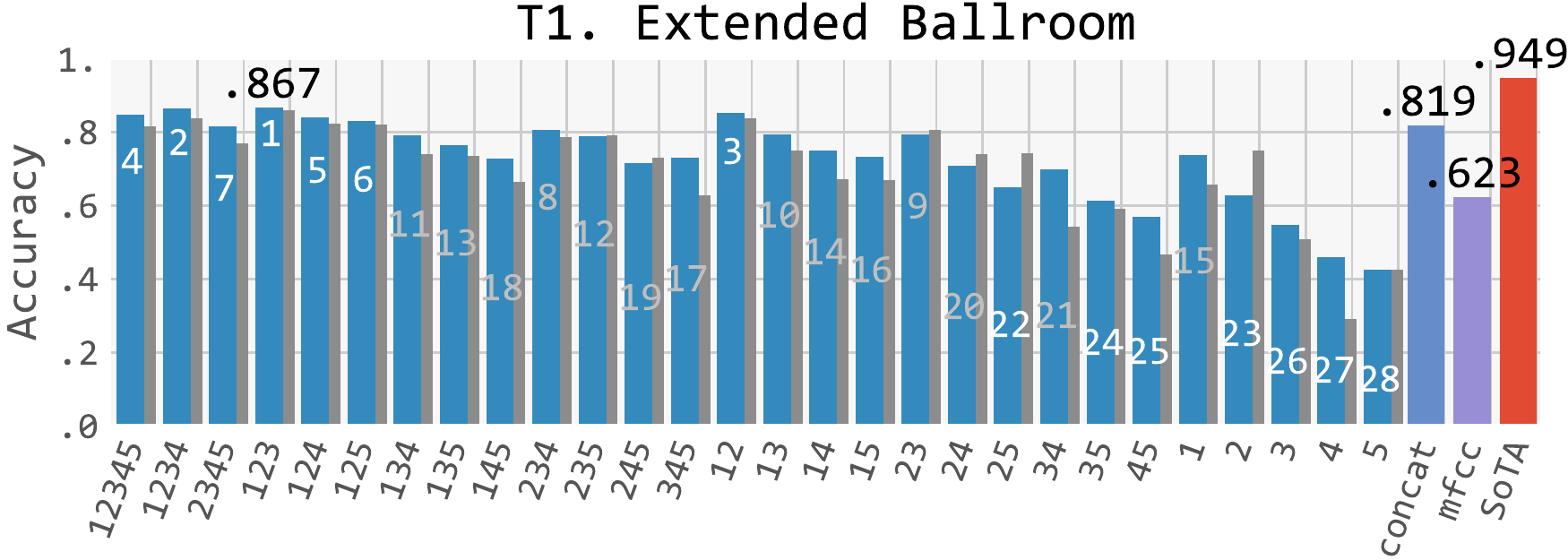}
	\vspace{-0.7cm}
	\caption{\small Performances of Task 1 - Ballroom dance genre classification of convnet features (with random convnet features in grey), MFCCs, and the reported state-of-the-art method.  (Note the exception that the SoTA is reported in weighted average recall.)} 
	\label{fig:result_all_t1}
%
	\vspace{0.2cm}
	\centering
	\includegraphics[width=1.0\columnwidth]{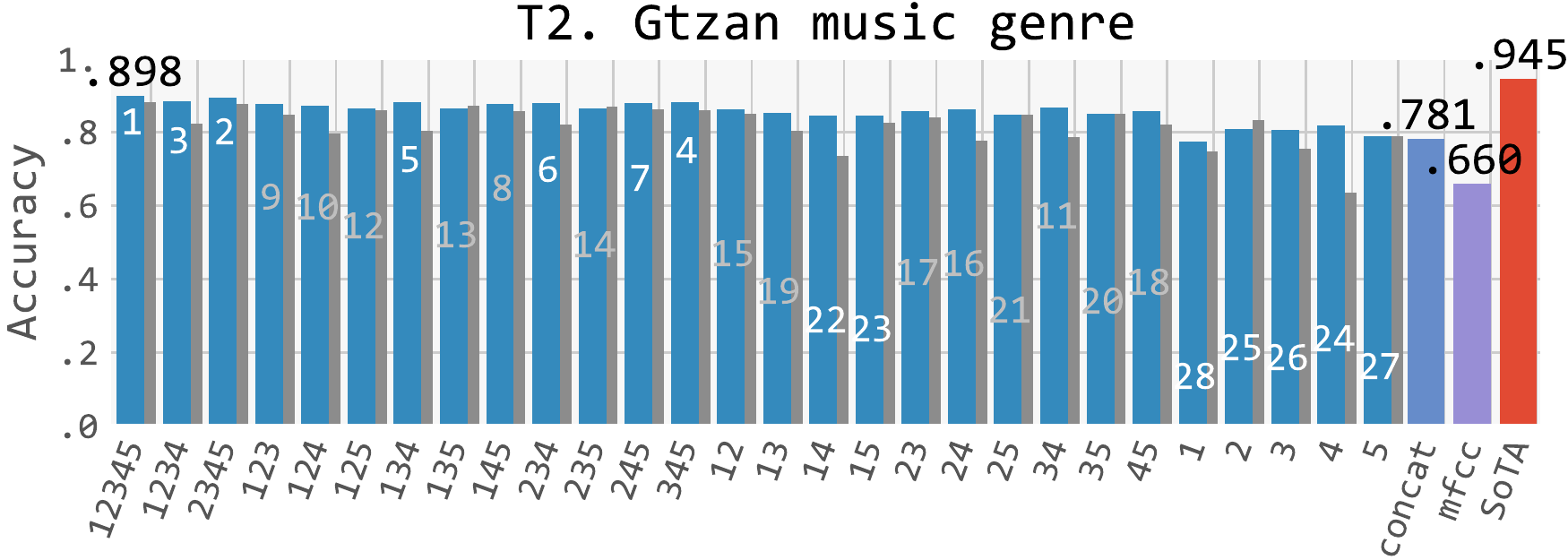}
	\vspace{-0.7cm}
	\caption{\small Performances of Task 2 - Gtzan music genre classification of convnet features (with random convnet features in grey), MFCCs, and the reported state-of-the-art method.}
	\label{fig:result_all_t2}
	\vspace{-0.2cm}
\end{figure}

Figure \ref{fig:result_all_t2} shows the performances on Gtzan music genre classification. The convnet feature shows 89.8\% while the concatenated feature and MFCCs respectively show only 78.1\% and 66.0\% of accuracy. Although there are methods that report accuracies higher than 94.5\%, we set 94.5\% as the state-of-the-art score following the dataset analysis in \cite{sturm2013gtzan}, which shows that the perfect score cannot surpass 94.5\% considering the noise in the Gtzan dataset.

Among a significant number of works that use the Gtzan music genre dataset, we describe four methods in more detail. Three of them use an SVM classifier, which enables us to focus on the comparison with our feature. Arabi~and~Lu~\cite{arabi2009enhanced} is most similar to the proposed convnet features in a way that it combines low-level and high-level features and shows a similar performance. Beniya~et~al.~\cite{baniya2014audio} and Huang~et~al.~\cite{huang2014music} report the performances with many low-level features before and after applying feature selection algorithms. Only the latter outperforms the proposed method and only after feature selection. 

\vspace{-0.2cm}
\begin{itemize}[leftmargin=*]
	\item Arabi~and~Lu~\cite{arabi2009enhanced} uses not only low-level features such as \{{spectral centroid/flatness/roll-off/flux}\}, but also high-level musical features such as \{{beat, chord distribution} and {chord progressions}\}. The best combination of the features shows 90.79\% of accuracy.
	\vspace{-0.2cm}
	\item Beniya~et~al.~\cite{baniya2014audio} uses a particularly rich set of statistics such as  \{{mean, standard deviation, skewness, kurtosis, covariance}\} of many low-level features including \{{RMS energy, attack, tempo, spectral features, zero-crossing, MFCC, dMFCC, ddMFCC, chromagram peak and centroid}\}. The feature vector dimensionality is reduced by {MRMR} (max-relevance and min-redundancy) \cite{peng2005feature} to obtain the highest classification accuracy of 87.9\%.	
	\vspace{-0.2cm}
	\item Huang~et~al.~\cite{huang2014music} adopts another feature selection algorithm, self-adaptive harmony search \cite{wang2010self}. The method uses statistics such as  \{{mean, standard deviation}\} of many features including \{{energy , pitch, and timbral features}\} and their derivatives. The original 256-dimensional feature achieved 84.3\% of accuracy which increases to 92.2\% and 97.2\% after feature selection.
	\vspace{-0.2cm}
	\item Reusing AlexNet \cite{krizhevsky2012imagenet}, a pre-trained convnet for visual image recognition achieved 78\% of accuracy \cite{gwardys2014deep}.
\end{itemize}
\vspace{-0.2cm}

In summary, the convnet feature achieves better performance than many approaches which use extensive music feature sets without feature selection as well as some of the approaches with feature selection.
For this task, it turns out that combining features from all layers is the best strategy. In the results, `\texttt{12345}', `\texttt{2345}', and `\texttt{1234}' are three best configurations, and all of the top-7 scores are from those strategies that use more than three layers. On the contrary, all lower-7 scores are from those with only 1 or 2 layers. This is interesting since the majority (7/10) of the target labels already exists in source task labels, by which it is reasonable to assume that the necessary information can be provided only with the last layer for those labels. Even in such a situation, however, low-level features contribute to improving the genre classification performance\footnote{On the contrary, in Task 5 - music emotion classification, high-level feature plays a dominant role (see Section \ref{sssec:exp_t4}).}.

\begin{figure}[t]
	\centering
	\centering
	\includegraphics[width=1.0\columnwidth]{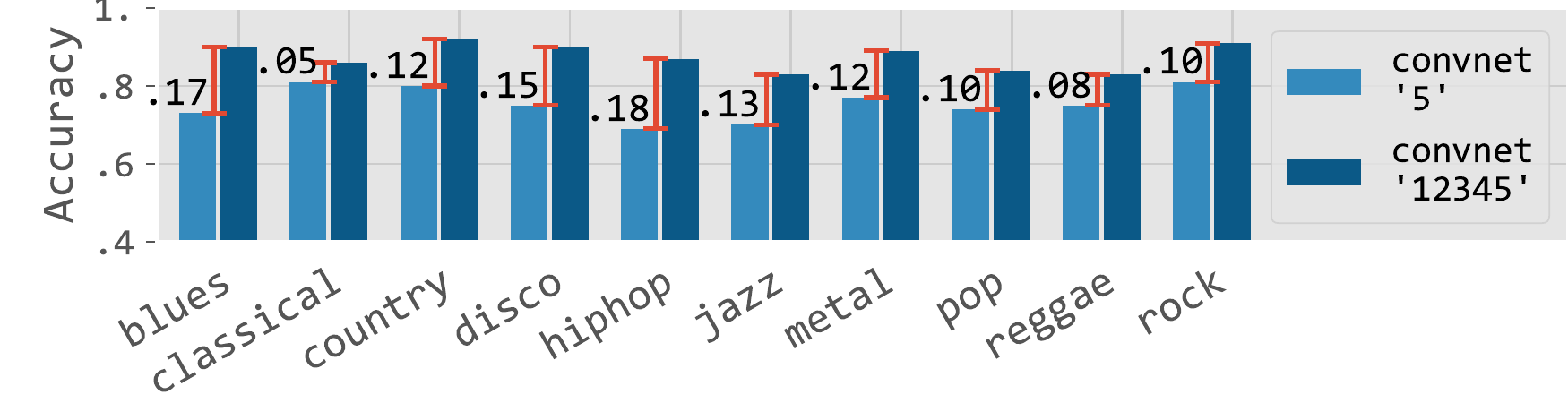}
	\vspace{-0.7cm}
	\caption{\small Comparison of per-label results of two convnet feature strategies, `\texttt{12345}' and `\texttt{5}' for Gtzan music genre classification. Numbers denote the differences of scores.}
	\label{fig:exp_gtzan_more}
%
\vspace{0.2cm}
	\centering
	\centering
	\includegraphics[width=1.0\columnwidth]{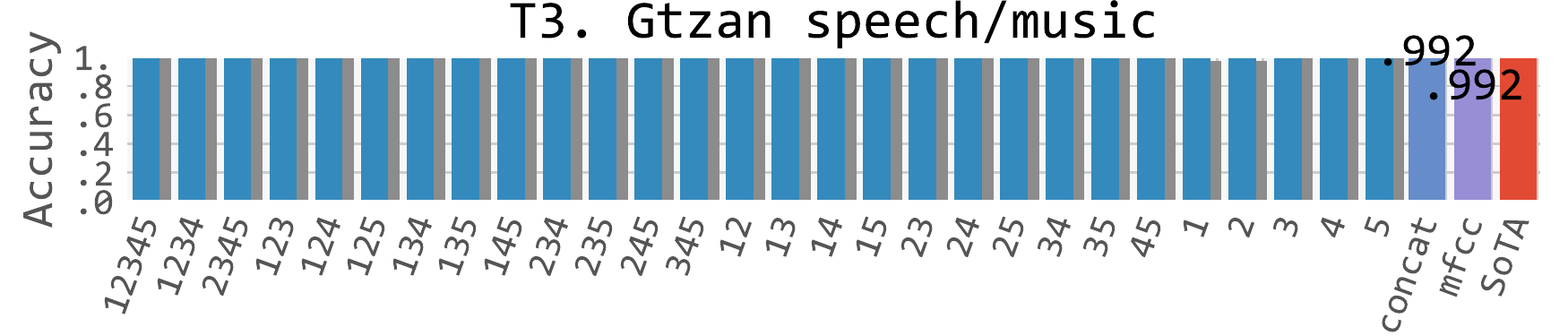}
	\vspace{-0.7cm}
	\caption{\small Performances of Task 3 - Speech/music classification of convnet features (with random convnet features in grey), MFCCs, and the reported state-of-the-art method. All scores of convnet features and SoTA are 1.0 and omitted in the plot.}
	\label{fig:result_all_t3}
\end{figure}

Among the classes of target task, \textit{classical} and \textit{disco}, \textit{reggae} do not exist in the source task classes. Based on this, we consider two hypotheses, \textit{i)} the performances of those three classes may be lower than the others, \textit{ii)} low-level features may play an important role to classify them since high-level feature from the last layer may be biased to the other 7 classes which exist in the source task.
However, both hypotheses are rebutted by comparing the performances for each genres with convnet feature `\texttt{5}' and `\texttt{12345}' as in Figure \ref{fig:exp_gtzan_more}. First, with `\texttt{5}' convnet feature, \textit{classical} shows the highest accuracy while both \textit{disco} and \textit{reggae} show accuracies around the average accuracy reported over the classes. Second, aggregating early-layer features affects all the classes rather than the three omitted classes. This suggests that the convnet features are not strongly biased towards the genres that are included in the source task and can be used generally for target tasks with music different from those genres.

\subsubsection{Task 3. Gtzan Speech/music Classification}

Figure \ref{fig:result_all_t3} shows the accuracies of convnet features, baseline feature, and state-of-the-art \cite{srinivas2014learning} with low-level features including MFCCs and sparse dictionary learning for Gtzan music/speech classification. A majority of the convnet feature combinations achieve 100\% accuracy. MFCC features achieve 99.2\%, but the error rate is trivial (0.8\% is one sample out of 128 excerpts).

Although the source task is only about music tags, the pre-trained feature in any layer easily solved the task, suggesting that the nature of music and speech signals in the dataset is highly distinctive.

\subsubsection{Task 4. Music Emotion Prediction} \label{sssec:exp_t4}

\begin{figure}[t]
	\centering
	\includegraphics[width=1.0\columnwidth]{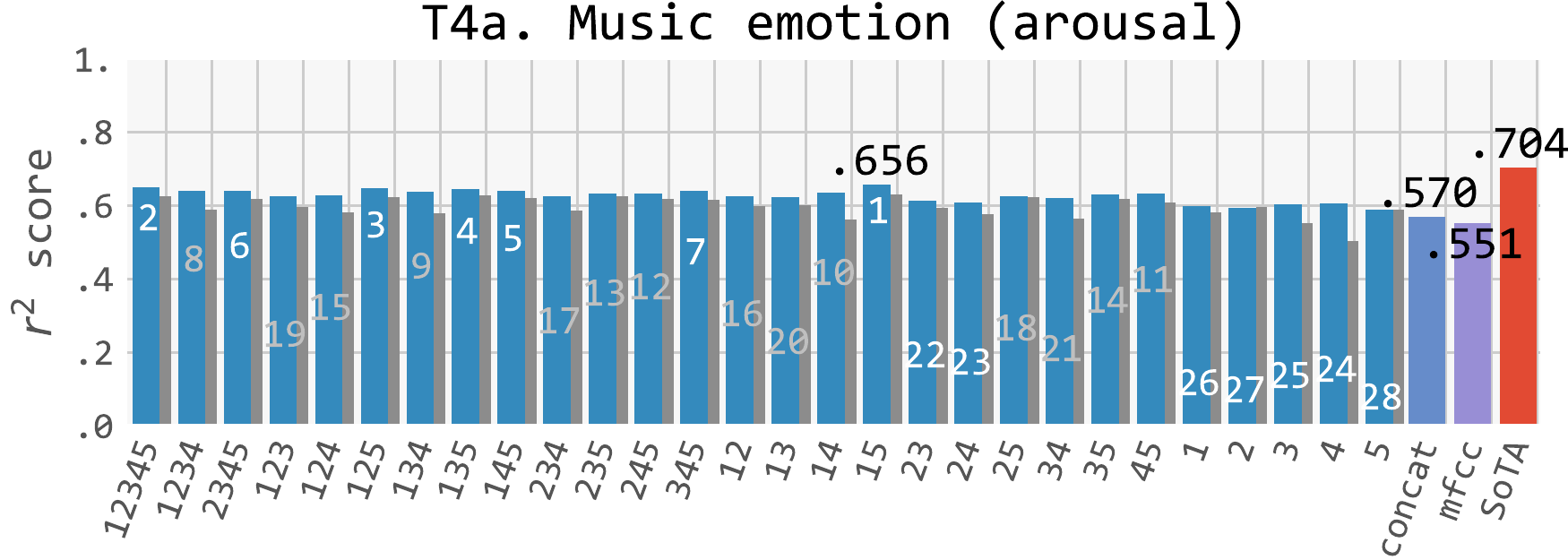}
	\newline		
	\includegraphics[width=1.0\columnwidth]{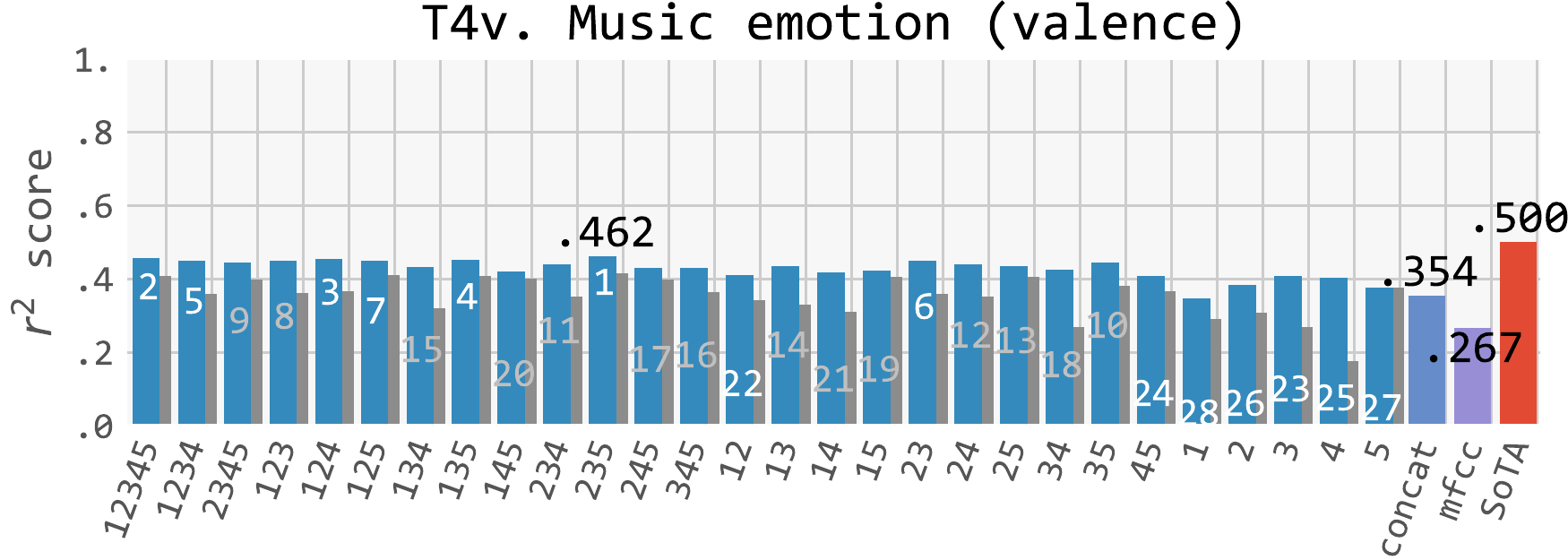}
\vspace{-0.7cm}
	\caption{\small Performances of Task 4a (arousal) and 4v (valence) - Music emotion prediction of convnet features (with random convnet features in grey), MFCCs, and the reported state-of-the-art method.}
	\label{fig:result_all_t4}
\end{figure}

Figure \ref{fig:result_all_t4} shows the results for music emotion prediction (Task 4). The best performing convnet features achieve 0.633 and 0.415 $r^2$ scores on arousal and valence axes respectively. 

On the other hand, the state-of-the-art algorithm reports 0.704 and 0.500 $r^2$ scores using music features with a recurrent neural network as a classifier \cite{weninger2014line} that uses 4,777 audio features including many functionals (such as \textit{quantiles, standard deviation, mean, inter peak distances}) of \textit{12 chroma features}, \textit{loudness}, \textit{RMS Energy}, \textit{zero crossing rate}, \textit{14 MFCCs}, \textit{spectral energy}, \textit{spectral roll-off}, etc.

For the prediction of arousal, there is a strong dependency on the last layer feature. All top-7 performances are from the feature vectors that include the fifth layer. The first layer feature also seems important, since all of the top-5 strategies include the first and fifth layer features.
For valence prediction, the third layer feature seems to be the most important one. The third layer is included in all of the top-6 strategies. Moreover, `\texttt{3}' strategy was found to be best performing among strategies with single layer feature.

To summarise the results, the predictions of arousal and valence rely on different layers, for which they should be optimised separately.
In order to remove the effect of the choice of a classifier and assess solely the effect of features, we compare our approach to the baseline method of \cite{weninger2014line} which is based on the same 4,777 features with SVM, not a recurrent neural network. 
The baseline method achieves .541 and .320 $r^2$ scores respectively on arousal and valence, both of which are lower than those achieved by using the proposed convnet feature. This further confirms the effectiveness of the proposed convnet features.

\subsubsection{Task 5. Vocal/non-vocal Classification}
\begin{figure}[t]
	\centering
	\centering
	\includegraphics[width=1.0\columnwidth]{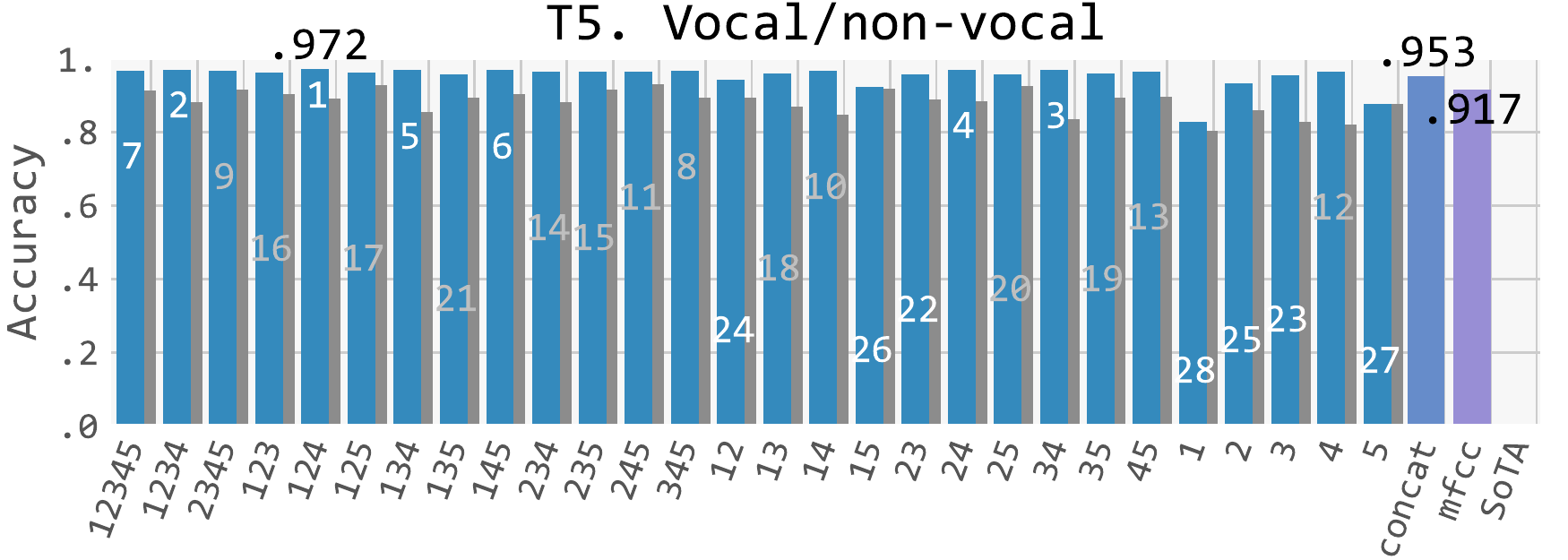}
	\vspace{-0.7cm}
	\caption{\small Performances of Task 5 - Vocal detection of convnet features (with random convnet features in grey) and MFCCs.}
	\label{fig:result_all_t5}
%
	\centering
	\centering
	\includegraphics[width=1.0\columnwidth]{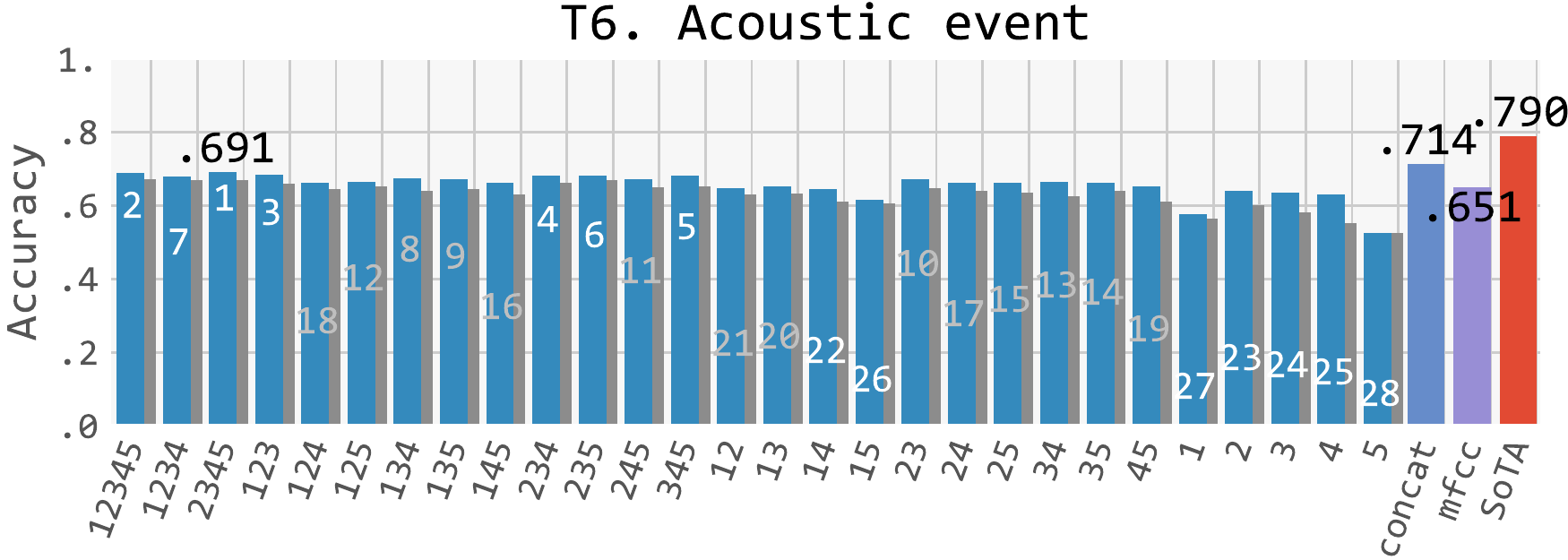}
	\vspace{-0.7cm}
	\caption{\small Performances of Task 6 - Acoustic event detection of convnet features (with random convnet features in grey), MFCCs, and the reported state-of-the-art method.}
	\label{fig:result_all_t6}
\end{figure}
\vspace{-0.2cm}

Figure \ref{fig:result_all_t5} presents the performances on vocal/non-vocal classification using the Jamendo dataset \cite{bibA03}. There is no known state-of-the-art result, as the dataset is usually used for \textit{frame-based} vocal \textit{detection/segmentation}. Pre-segmented \textit{Excerpt} \textit{classification} is the task we formulate in this paper. For this dataset, the fourth layer plays the most important role. All the 14 combinations that include the fourth layer outperformed the other 14 strategies without the fourth layer.

\subsubsection{Task 6. Acoustic Event Detection}

Figure \ref{fig:result_all_t6} shows the results on acoustic event classification using Urbansound8K dataset \cite{Salamon:UrbanSound:ACMMM:14}. Since this is not a music-related task, there are no common tags between the source and target tasks, and therefore the final-layer feature is not expected to be useful for the target task. 

The strategy of concatenating `\texttt{12345}' convnet features and MFCCs yields the best performance. Among convnet features, `\texttt{2345}', `\texttt{12345}', `\texttt{123}', and `\texttt{234}' achieve good accuracies. In contrast, those with only one or two layers do not perform well. We were not able to observe any particular dependency on a certain layer.

Since the convnet features are trained on music, they do not outperform a dedicated convnet trained for the target task. The state-of-the-art method is based on a deep convolutional neural network with data augmentation \cite{salamon2017deep}. Without augmenting the training data, the accuracy of convnet in the same work is reported to be 74\%, which is still higher than our best result (71.4\%).\footnote{Transfer learning targeting audio event classification was recently introduced in \cite{aytar2016soundnet, arandjelovic2017look} and achieved a state-of-the-art performance.}

The convnet feature still shows better results than conventional audio features, demonstrating its versatility even for non-musical tasks. The method in \cite{Salamon:UrbanSound:ACMMM:14} with \{\textit{minimum, maximum, median, mean, variance, skewness, kurtosis}\} of 25 MFCCs and \{\textit{mean and variance}\} of the first and second MFCC derivatives (225-dimensional feature) achieved only 68\% accuracy using the SVM classifier. This is worse than the performance of the best performing convnet feature.

It is notable again that unlike in the other tasks, concatenating convnet feature and MFCCs results in an improvement over either a convnet feature or MFCCs (71.4\%). This suggests that they are complementary to each other in this task.

\section{Conclusions}

We proposed a transfer learning approach using deep learning and evaluated it on six music information retrieval and audio-related tasks. The pre-trained convnet was first trained to predict music tags and then aggregated features from the layers were transferred to solve genre classification, vocal/non-vocal classification, emotion prediction, speech/music classification, and acoustic event classification problems. Unlike the common approach in transfer learning, we proposed to use the features from every convolutional layers after applying an average-pooling to reduce their feature map sizes.

In the experiments, the pre-trained convnet feature showed good performance overall. It outperformed the baseline MFCC feature for all the six tasks, a feature that is very popular in music information retrieval tasks because it gives reasonable baseline performance in many tasks. It also outperformed the random-weights convnet features for all the six tasks, demonstrating the improvement by pre-training on a source task. Somewhat surprisingly, the performance of the convnet feature is also very competitive with state-of-the-art methods designed specifically for each task. The most important layer turns out to differ from task to task, but concatenating features from all the layers generally worked well. For all the five \textit{music} tasks, concatenating MFCC feature onto convnet features did not improve the performance, indicating the music information in MFCC feature is already included in the convnet feature. We believe that transfer learning can alleviate the data sparsity problem in MIR and can be used for a large number of different tasks.

\bibliography{music_transferred_learning}

\end{document}